\documentclass[conference]{IEEEtran}
\IEEEoverridecommandlockouts
\usepackage{cite}
\usepackage{amsmath,amssymb,amsfonts}
\usepackage{algorithmic}
\usepackage{graphicx}
\usepackage{textcomp}
\usepackage{xcolor}

\usepackage{tcolorbox}
\usepackage{bm}
\usepackage{booktabs}
\usepackage{multirow}
\usepackage{makecell}

\def\BibTeX{{\rm B\kern-.05em{\sc i\kern-.025em b}\kern-.08em
    T\kern-.1667em\lower.7ex\hbox{E}\kern-.125emX}}
\begin{document}

% 论文名称
\title{Towards Aligned Data Forgetting via Twin Machine Unlearning}

%\author{
%    \IEEEauthorblockN{
%        Zhenxing Niu\textsuperscript{1}, 
%        Haoxuan Ji\textsuperscript{2}, 
%        Yuyao Sun\textsuperscript{1}, 
%        Zheng Lin\textsuperscript{1}, 
%        Fei Gao\textsuperscript{1}, 
%        Yuhang Wang\textsuperscript{1}, 
%        Haichao Gao\textsuperscript{1}
%    }
%    \IEEEauthorblockA{
%        \textsuperscript{1}Xidian University, Xian, China\\
%        \textsuperscript{2}Xi'an Jiaotong University, Xian, China\\
%        Emails: zhenxingniu@gmail.com, xjtujhx@stu.xjtu.edu.cn, yysun@stu.xidian.edu.cn, \\
%        linzheng@stu.xidian.edu.cn, fgao@xidian.edu.cn, 24031212230@stu.xidian.edu.cn, hchgao@xidian.edu.cn
%    }
%}

% 双盲，匿名
\author{ICME submission 1754}

\author{

\IEEEauthorblockN{Haoxuan Ji}
\IEEEauthorblockA{\textit{Xi'an Jiaotong University} \\
Xian, China \\
xjtujhx@stu.xjtu.edu.cn}\\

\and

\IEEEauthorblockN{Zheng Lin}
\IEEEauthorblockA{\textit{Xidian University} \\
Xian, China \\
24031212230@stu.xidian.edu.cn}

\and

\IEEEauthorblockN{Yuyao Sun}
\IEEEauthorblockA{\textit{Xidian University} \\
Xian, China \\
22031212453@stu.xidian.edu.cn}

\and

\IEEEauthorblockN{Gao Fei}
\IEEEauthorblockA{\textit{Xidian University} \\
Xian, China \\
fgao@xidian.edu.cn}\\

\and

\IEEEauthorblockN{Yuhang Wang}
\IEEEauthorblockA{\textit{Xidian University}\\
Xian, China \\
24031212230@stu.xidian.edu.cn}

\and

\IEEEauthorblockN{Haichang Gao}
\IEEEauthorblockA{\textit{Xidian University} \\
Xian, China \\
hchgao@xidian.edu.cn}\\

\and

\IEEEauthorblockN{Zhenxing Niu}
\IEEEauthorblockA{\textit{Xidian University} \\
Xian, China \\
zxniu@xidian.edu.cn}\\
}

\maketitle

% 摘要
\begin{abstract}
Modern privacy regulations have spurred the evolution of machine unlearning, a technique enabling a trained model to efficiently forget specific training data. In prior unlearning methods, the concept of ``data forgetting" is often interpreted and implemented as achieving \emph{zero} classification accuracy on such data. Nevertheless, the authentic aim of machine unlearning is to achieve \emph{alignment} between the unlearned model and the gold model, \emph{i.e.}, encouraging them to have \emph{identical} classification accuracy. On the other hand, the gold model often exhibits \emph{non-zero} classification accuracy due to its generalization ability. To achieve aligned data forgetting, we propose a Twin Machine Unlearning (TMU) approach, where a twin unlearning problem is defined corresponding to the original unlearning problem. Consequently, the generalization-label predictor trained on the twin problem can be transferred to the original problem, facilitating \emph{aligned data forgetting}. Comprehensive empirical experiments illustrate that our approach significantly enhances the alignment between the unlearned model and the gold model.
\end{abstract}
%关键字
% 介绍
\section{Introduction}
Machine learning model providers tend to collect extensive data from the Internet and utilize it to train their machine learning models. The recent introduction of data privacy and protection regulations (European Union’s GDPR \cite{regulation2018general}, and California Consumer Privacy Act (CCPA) \cite{goldman2020introduction}) obligate these model providers to comply with the request-to-delete \cite{dang2021right} from the data owner. For example, a corporation offering facial recognition services might acquire facial images from the Internet to train their facial recognition models. Subsequently, a user might discover that the company has utilized his/her facial images in model training. In such a scenario, the user reserves the right to petition the company to revise the facial recognition model to forget his/her facial image data.

The straightforward solution is to entirely discard the trained model, delete his/her facial images from the training data, and re-train a new model from scratch (\emph{i.e.,} called \emph{gold model}). Unfortunately, re-training from scratch is expensive. Therefore, machine unlearning \cite{garg2020formalizing,ginart2019making,cao2015towards, gupta2021adaptive, sekhari2021remember, brophy2021machine,ullah2021machine} has garnered considerable attention, aiming to \emph{efficiently revise} a trained model to forget a cohort of training data, without affecting the performance on the remaining data.

In most prior unlearning methods \cite{graves2021amnesiac, tarun2023fast}, the ``forgetting'' of a cohort data $D_f$ is commonly interpreted and implemented by decreasing the classification accuracy on $D_f$ \emph{as much as possible}, \emph{i.e.,} pursuing \emph{zero} accuracy $ACC^{unlearn}_{D_f}=0$.
This implementation works well when forgetting an \textbf{entire} class $k$ (\emph{i.e.,} $D_f=D_k$). However, it is not suitable when the aim is to forget only \textbf{a subset} of class $k$ (\emph{i.e.,} $D_f \subset D_k$). In this case, even though $D_f$ are not involved in the training of the gold model, the remaining samples from class $k$ (\emph{i.e.,} samples $\in (D_k - D_f)$) are still involved in training. Thus, the gold model has ability to recognize class $k$.
Consequently, the gold model can still correctly classify a portion of $D_f$ due to its generalization ability, resulting in $ACC^{gold}_{D_f} \neq 0$. As we know, the actual aim of machine unlearning is to obtain an unlearned model that is well \emph{aligned} with the gold model. Here, ``align" implies \textbf{requiring the unlearned model to possess the \emph{identical} classification accuracy as the gold model}. Obviously, it is not appropriate to pursue $ACC^{unlearn}_{D_f}=0$. 

\iffalse
\begin{figure*}[t]
\centering
%\vspace{-1.0em}
% \setlength{\abovecaptionskip}{0.4cm}
\includegraphics[width=0.8\linewidth]{figs/fig1.png}
\caption{Construction of the \emph{Twin Model} and corresponding \emph{Twin Unlearning Problem}.}\label{fig1}
\vspace{-2.0em}
\end{figure*}
\fi

In this paper, we denote the samples $x\in D_f$ that can be correctly classified by the gold model as the `\emph{easy}' samples. Conversely, the samples that cannot be correctly classified are referred to as the `\emph{hard}' samples.  Therefore, a \emph{well-aligned} unlearning algorithm should only decrease the classification accuracy on \emph{hard} samples rather than \emph{all} samples in $D_f$. In other words, it is essential to first identify the hard samples from $D_f$, and subsequently reduce the classification accuracy on such samples.
However, since the gold model is unknown, distinguishing between easy and hard samples becomes challenging. This is the reason why prior unlearning methods have to adopt achieving \emph{zero} classification accuracy as the objective of data forgetting.

In this paper, we aim to achieve aligned data forgetting. As these easy/hard labels for $D_f$ are defined based on the gold model's generalization ability, we refer to them as \emph{generalization-labels}. In our approach, we propose to train a binary classifier to predict the generalization-labels for $D_f$. With these predicted labels, $D_f$ is partitioned into an easy subset $D^e_f$ and a hard subset $D^h_f$. Subsequently, we specifically decrease the classification accuracy on $D^h_f$, while retaining the classification accuracy on $D^e_f$. The most challenging aspect of our approach lies in obtaining such a binary classifier, which involves two sub-challenges: (1) build a specific labeled dataset for training the binary classifier; (2) construct discriminative features for the binary classifier.

\begin{figure*}[t]
\centering
\includegraphics[width=0.7\linewidth]{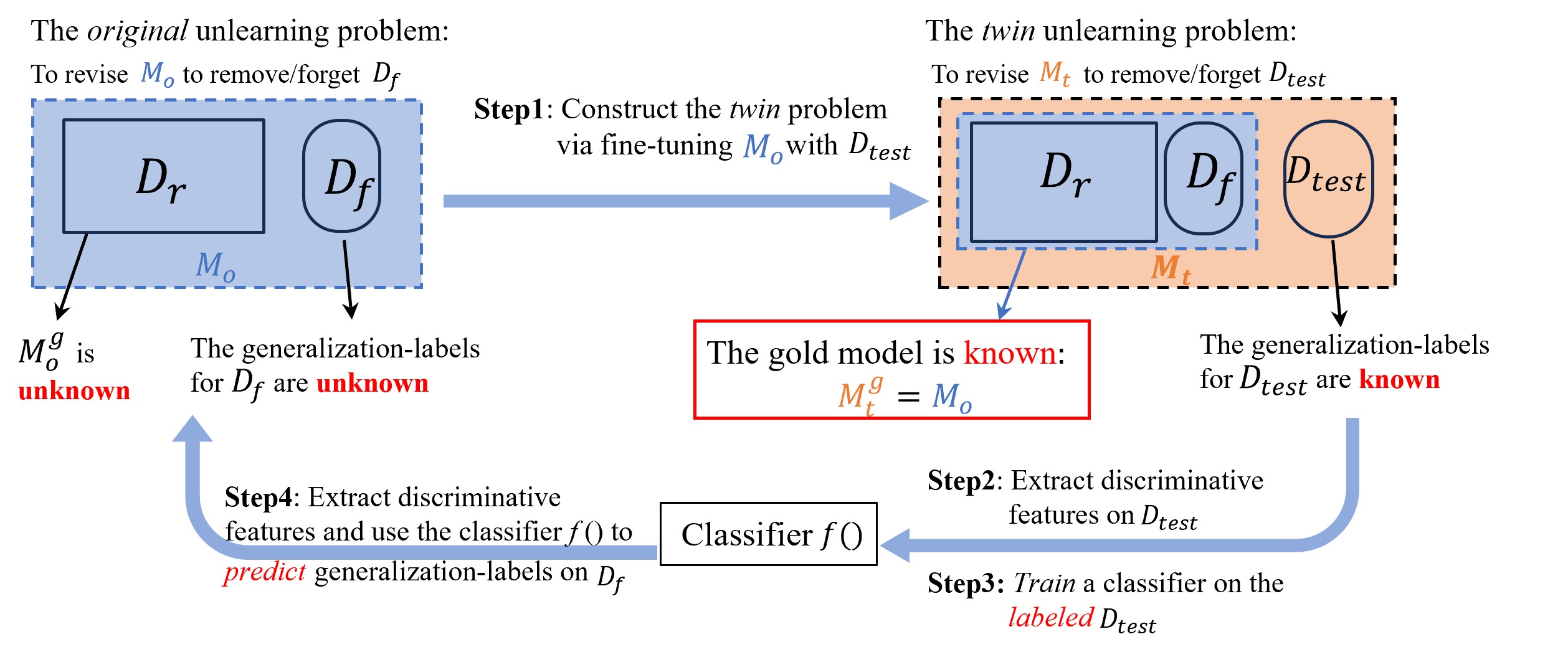}
\vspace{-1.0em}
\caption{The workflow of our approach. We first construct the twin model by fine-tuning the $M_o$ with $D_{test}$ to produce $M_t$. And then, we extract discriminative features and train a binary classifier on $D_{test}$. Consequently, we transfer the binary classifier to the original problem to predict the generalization-labels on $D_f$. Finally, we reduce classification accuracy on $D_f^h$.}\label{fig2}
\vspace{-1.0em}
\end{figure*}

To address the first challenge, we formulate a twin unlearning problem corresponding to the original one. Specifically, we first fine-tune the original model $M_{o}$ using an additional dataset $D_{test}$, resulting in a new model $M_{t}$, referred to as the \emph{Twin Model} in this paper. In the context of $M_{t}$, the original model $M_{o}$ can be seen as its gold model, meanwhile $D_{test}$ can be seen as its forgetting data. In another word, given the twin model $M_{t}$, the \emph{Twin Unlearning Problem} is defined as the process of enabling $M_{t}$ to forget $D_{test}$. Note that in the twin unlearning problem, the gold model is known $M_t^g=M_{o}$. 

The reason for formulating such a twin unlearning problem is that we can easily obtain a labeled dataset -- the dataset $D_{test}$. Since the gold model of the twin unlearning problem is known (\emph{i.e.,} $M_{o}$), we can easily obtain the generalization-labels for $D_{test}$, specifically by employing $M_{o}$ to make class predications on $D_{test}$.

Regarding the second challenge, we propose to integrate three complementary features, each possessing strong discriminability to distinguish between easy and hard samples. (1) The first feature is the Distance Feature (DF). For each sample $x$ in $D_f$, we identify its nearest neighbor in $D_r$, and the distance between them is considered as a feature. Intuitively, if $x$ is an easy sample, it typically has similar samples in $D_r$, resulting in a small distance. In contrast, the distance feature for a hard sample tends to be large. (2) The second feature is the Adversarial-attack Feature (AF). We construct this feature by employing adversarial attacks\cite{madry2017towards}. Clearly, hard samples are more vulnerable to adversarial attacks compared to easy samples due to their proximity to the decision boundary. Hence, the results of adversarial attack can be considered a discriminative feature.(3) The third feature is the Curriculum-learning-loss Feature (CF). Curriculum learning theory \cite{kumar2010self,bengio2009curriculum} suggests that easy samples are learned earlier than hard samples. Therefore, the loss value of the first fine-tuning epoch is considered a feature to distinguish between easy and hard samples.

\iffalse
Overall, we summarize our contributions as follows.
\begin{itemize}
    \item [$\bullet$] We emphasize alignments in data forgetting and introduce a Twin Machine Unlearning (TMU) approach to enhance such alignments. 		
    \item [$\bullet$] We devise three discriminative features to distinguish easy and hard samples, employing adversarial attack and curriculum learning strategies.
    \item [$\bullet$] To withstand the assessment of data forgetting from MIA, we propose a noise-perturbed fine-tuning scheme to enhance model's resilience to MIA. 
\end{itemize}
\fi

\section{Problem of Data Forgetting}
Let $D=\{x_i, y_i\}^N_{i=1}$ be a dataset of images $x_i$, each with a label $y_i\in \{1,\dots K\}$ representing a class. The original model $M_{o}(\theta)$ is trained with the $D$, which could be a DNN with parameters $\theta$.
Let $D_f$ be a subset of the $D$, whose information we want to forget from the trained model $M_{o}(\theta)$. The $D_f$ is called the \emph{forgetting data}, which could be \emph{all} or \emph{some} of the data with a given label $k$, \emph{i.e.,} corresponding to forgetting an entire class or a subset of a class, respectively. In this paper, we are particularly interested in the case of forgetting  a subset of a class, as the alignment problem becomes more critical in this scenario. The \emph{remaining data} is denoted by $D_r$, whose information is desired to be kept unchanged in the model. $D_f$ and $D_r$ together represent the entire training set $D$ and are mutually exclusive, \emph{i.e.,} $D_r \cup D_f =D$ and $D_r \cap D_f =\emptyset$.

Since $D_f$ has been used to train the $M_{o}(\theta)$, the model parameters $\theta$ will contain information about $D_f$. The task of data forgetting aims to \textbf{forget the information of $D_f$ from the trained model $M_{o}(\theta)$}. The ideal solution is to train a new model from scratch with $D_r$, which is denoted as the \emph{gold} model $M_{o}^g(\theta_r)$. However, obtaining the gold model is time-consuming due to the expensive re-training procedure. Thus, the practical data forgetting aims to efficiently revise the original model $\theta$ with a `scrubbing/forgetting' function $s( )$, so that the revised model $s(\theta)$ is as close to the gold model $\theta_r$ as possible. 
The $s( )$ is often implemented by a \emph{machine unlearning} algorithm. 

\noindent\textbf{Alignment in Data Forgetting}
The model alignment is often measured by some metrics (called \emph{readout functions} \cite{golatkar2021mixed}), such as: (i) accuracy on the test set $D_{test}$, \emph{i.e.,} $ACC_{D_{test}}$; (ii) accuracy on the forgetting data $D_f$, \emph{i.e.,} $ACC_{D_f}$; (iii) accuracy on the remaining data $D_r$, \emph{i.e.,} $ACC_{D_r}$. Achieving alignment on $ACC_{D_{test}}$ and $ACC_{D_r}$ is straightforward, as the target is to increase them as much as possible. In contrast, the challenge lies in aligning the $ACC_{D_f}$, \emph{i.e.,} the unlearned model is desired to have the same classification accuracy as the gold model $ACC^{unlearn}_{D_f}=ACC^{gold}_{D_f}$. Since the gold model is unknown, obtaining its $ACC^{gold}_{D_f}$ is impossible, let alone achieving accuracy alignment. 

As a result, prior unlearning methods tend to replace this objective with a surrogate objective, \emph{i.e.,} decreasing the accuracy $ACC_{D_f}$ as much as possible. For the case of forgetting an entire class, this implementation is acceptable since, after removing an entire class, the gold model tends to have $ACC^{gold}_{D_f}=0$, aligning with the surrogate objective. However, in the case of forgetting a subset of class, the remaining data will still contain samples of the forgetting class. Due to the generalization ability of gold model, the gold model may correctly classify some samples in $D_f$, \emph{i.e.,} $ACC^{gold}_{D_f} \neq 0$. Thus, we have to carefully consider the alignment challenges.

\section{Our Approach}
In our approach, we first construct a new unlearning problem corresponding to the original one. As shown in the left part of Fig.\ref{fig2}, for the original unlearning problem, we have an \emph{original model} $M_{o}$ that is trained with $D_r$+$D_f$. Our task is to get an \emph{unlearned model} $M_{u}$ from $M_{o}$, so that $M_{u}$ is well aligned with the \emph{gold model} $M_{o}^g$. The $M_{o}^g$ is a model trained from scratch only with $D_r$.

\subsection{Twin Unlearning Problem}

As shown in the right part of Fig.\ref{fig2}, if we fine-tune the $M_{o}$ with another dataset $D_{test}$, we obtain a new model, namely the twin model $M_{t}$. Consequently, we can construct a Twin Unlearning Problem: we regard the $M_{t}$ as the ``original model'', which is trained with $D_r$+$D_f$+$D_{test}$. We regard the $D_{test}$ as the ``forgetting data''. 
Thus, our twin unlearning problem is defined as:
\begin{tcolorbox}
    Given a trained model $M_{t}$, the unlearning task is to forget $D_{test}$ from $M_{t}$.
\end{tcolorbox}

Obviously, the twin problem's ``gold model'' $M_{t}^g$ is known, which is exactly the original problem's original model $M_o$,
\begin{align}
    M_{t}^g = M_{o} 
\end{align}
Regarding the constructed twin unlearning problem, since its gold model $M_{t}^g$ is known, we can easily obtain the ground-truth generalization-labels for its forgetting data $D_{test}$, \emph{i.e.,} run the inference of the model $M_{t}^g$ over the $D_{test}$ to obtaining the classification results. So far, \textbf{we have build a specific labeled dataset.} We can train a binary classifier $f()$ to distinguish between hard and easy samples on $D_{test}$ for the twin problem. 

It is usually to assume that $D_{test}$ and $D_f$ have independent and identical distribution (I.I.D). Thus, the binary classifier $f()$ can be transferred from the twin problem back to the original unlearning problem for predicting the generalization-labels on $D_f$. In this way, we can effectively partition $D_f$ into a hard subset $D_f^h$ and an easy subset $D_f^e$. Finally, when solving the original unlearning problem, we can fine-tune the original model $M_{o}$ by reducing classification accuracy on $D_f^h$ (\emph{i.e.,} maximizing its loss value) while retaining the accuracy on $D_f^e$ (\emph{i.e.,} maximizing its negative loss value), as follows:
\begin{align}
    &\mathop{\max}_{\theta} (L_{D_f^h}(x,y;\theta) - L_{D_f^e}(x,y;\theta))\label{eq:fh0}  
\end{align}

\subsection{Discriminative Features}

Besides building a specific labeled dataset for training the binary classifier, another core of our approach is \textbf{to construct discriminative features for the classifier}. Briefly, we propose to integrate three complementary features by employing adversarial attack and curriculum learning strategies. 

%\subsubsection{Distance Feature (DF)}
\noindent\textbf{Nearest-distance Feature:}
In the case of forgetting a subset of a class $k$, the remaining data $D_r$ will still contain samples belonging to the forgetting class $k$. Therefore, for each sample $x\in D_f$, we can find some samples $x'\in D_r$ similar to $x$. The similarity can be measured by the distance $l(x,x')$ with respect to a feature extractor $g()$, 
\begin{align}
    l(x,x')=|g(x) - g(x')| 
\end{align}

For each $x\in D_f$, we can identify the top-$N$ similar samples $x_i'\in D_r, (i=0,\dots,N-1)$. Intuitively, these distances $l(x,x_i')$ for an easy sample would be smaller than that for a hard sample. Thus, the concatenation of these top-$N$ distances $l(x,x_i'), (i=0,\dots,N-1)$ is regarded as a discriminative feature to distinguish between easy and hard samples. 

%\subsubsection{Adversarial-attacking Feature (AF)}
\noindent\textbf{Adversarial-attacking Feature:}
Whether a sample can be correctly generalized or not (\emph{i.e.,} belong to a easy or hard sample) is closely related to the positional relation between it and the classification boundary. If it is near the boundary, it is hard to correctly classify it with high confidence. Thus, the easy samples often stay far from the boundary, while the hard samples tend to stay near the boundary.

In order to leverage such positional relation knowledge to distinguish between easy and hard samples, we employ the untargeted adversarial attack technique. As we know, given any sample $x$, we can generate a corresponding adversarial sample $\tilde{x}$ whose classification result is different from its ground-truth label. The adversarial sample $\tilde{x}$ is often generated by perturbing $x$ with a certain perturbation $r$. The perturbation $r$ is typically constrained by a certain attack budget $\epsilon$, as follows,
\begin{align}
    &\mathop{\max}_{\bm{r}} L(\widetilde{\bm{x}},y;\theta) \label{eq:adv} \\
    &\text{s.t. }  ||\bm{r}||_p < \epsilon, \widetilde{\bm{x}}=\bm{x+r} \notag
\end{align}
The essence of adversarial attack is to move $x$ across the classification boundary, \emph{i.e.,} changing the classification result. Thus, given a sufficiently large budget $\epsilon$, we can successfully conduct the adversarial attack on any sample, \emph{i.e.,} successfully generating $\tilde{x}$.

Intuitively, the attack budget for a sample is related to the positional relation between it and the classification boundary. If the sample is near the boundary, a small budget is enough to conduct a successful attack. Thus, we can distinguish between hard and easy samples like that: we use \emph{a relatively small attack budget} to conduct adversarial attack, and we identify easy and hard samples based on whether the attack is successful or fails, \emph{i.e.,} hard samples can be successfully attacked with a small attack budget, since it is near the classification boundary.

Furthermore, the bipolar results about successful or failure attack cannot be regarded as a good continuous discriminative feature. Instead, we compute the classification score/logits $s(\tilde{x})$ for the adversarial sample $\tilde{x}$. On the other hand, we calculate the classification score $s(x)$ for the clean sample $x$, and measure the cross-entropy between $s(\tilde{x})$ and $s(x)$,
\begin{align}
    dist(x, \tilde{x}) = H(s(\tilde{x}), s(x)) 
\end{align}
Obviously, the $dist(x, \tilde{x})$ will be large if the attack is successful and vice versa. The larger the $dist(x, \tilde{x})$ is, the higher the likelihood of a successful attack. Thus, the $dist(x, \tilde{x})$ is regarded as the second discriminative feature to distinguish between easy and hard samples.

\noindent\textbf{Curriculum-learning-loss Feature:}
Curriculum learning theory suggests that easy samples are learned earlier than hard samples. 
Thus, we propose to employ a curriculum learning strategy to build our third feature. 
Curriculum learning has illustrated that neural networks tend to learn easy samples very quickly at the beginning of the training iterations. Conversely, the hard samples are learned at the later iterations. Specifically, we train a \emph{randomly-initialized} network $M_r$ from scratch, where the architecture of $M_r$ is the same as the architecture of original model $M_o$. To distinguish between easy and hard samples, we just train the model $M_r$ for \emph{one} or \emph{two} epochs, instead of completing the full training procedure.

Regarding whether a sample $x$ has been well-learned by $M_r$, we use the loss $loss(x)$ as a metric, \emph{i.e.,} a small loss value implies that $x$ has been well learned (indicating it is a easy sample) while a large loss value implies that $x$ has not been well learned (indicating it is a hard sample). Therefore, the curriculum-learning-loss $loss(x)$ is considered as the third feature to distinguish between easy and hard samples.

Note that the model $M_r$ is trained with a part of $D_r+D_f$ and $D_{test}$. In practice, we find that it is enough to train $M_r$ by just utilizing $30\%$ of all $D_r+D_f$. Thus, the training of $M_r$ is much cheaper than the training of $M_o$.

\subsection{Binary Classifier}
We have build a specific labeled dataset $D_{test}$. On the other hand, we have devised three discriminative features to distinguish between easy and hard samples. Next, we will train a binary classifier to distinguish between easy and hard samples with these feature over the labeled dataset. Specifically, we employ an MLP network with two hidden layers as the binary classifier. We concatenate the three discriminative features as the input of the binary classifier. 
Note that we have opted for a straightforward binary classifier due to the discriminative nature of our features and our desire to keep the unlearning process cost-effective.

\begin{table}[tpb]
\caption{Comparison with four unlearning methods on CIFAR-10. We conduct the data forgetting experiment for each class in CIFAR-10 independently. The results in this table are the \textbf{average values} over all 10 classes in CIFAR-10.}
\scriptsize
\centering
\label{defense}
\begin{tabular}{cccccc}
\toprule
    \multicolumn{2}{c}{\makecell{Experiment}}& \multicolumn{3}{c}{\makecell{Accuracy on $D_{test}$ and $D_{f}$}} & \makecell{Activation\\Distance} \\
    \cmidrule(lr){1-2} \cmidrule(lr){3-5}  \cmidrule(lr){6-6}  
    Model& Methods &$ACC_{D_{test}}$$\uparrow$ & $ACC_{D_{f}}$ & $\Delta$$\downarrow$ & AD$\downarrow$ \\       
\midrule
\multirow{6}{*}{\makecell{ResNet-18}}&Gold Model         &85.31 &86.9 &0     & 0      \\
&Fine-tuning     &85.06 &100  &13.1  & 0.58   \\
&NegGrad &84.81 &10.6 &76.3  & 0.72   \\
&RandLabel    &85.43 &0    &86.9  & 1.25   \\
&Bad Teacher     &82.37 &91.5 &8.0   & 0.71   \\
&Ours            &84.86 &90.8 &4.2   & 0.49   \\
\midrule
\multirow{6}{*}{\makecell{ALLCNN}}&Gold Model         &86.58 &86.4 &0     & 0      \\
&Fine-tuning     &87.34 &100.0 &13.1  & 0.32  \\
&NegGrad &86.87 &3.3   &84.3  & 0.84  \\
&RandLabel    &86.89 &0     &86.4  & 0.96  \\
&Bad Teacher     &85.96 &88.0  &5.8   & 0.49  \\
&Ours            &86.78 &87.7  &4.9   & 0.29  \\
\midrule
\multirow{6}{*}{\makecell{ViT}}&Gold Model         &84.61 &84.6 &0     & 0      \\
&Fine-tuning     &84.83 &100.0 &15.4  & 0.65  \\
&Neg. Gradient&84.51 &62.7  &23.9  & 0.92  \\
&Rand. Label    &84.21 &2.4   &82.2  & 0.88  \\
&Bad Teacher     &83.05 &87.9  &5.3   & 0.59  \\
&Ours            &84.07 &86.1  &3.3   & 0.40  \\

\bottomrule
\end{tabular}
\label{main_result}
\end{table}

\begin{table}[h]
\caption{Comparison with four unlearning methods on CIFAR-100.}
\scriptsize
\centering
\label{defense}
\begin{tabular}{cccccc}
\toprule
    \multicolumn{2}{c}{\makecell{Experiment}}& \multicolumn{3}{c}{\makecell{Accuracy on $D_{test}$ and $D_{f}$}} & \makecell{Activation\\Distance}  \\
    \cmidrule(lr){1-2} \cmidrule(lr){3-5}  \cmidrule(lr){6-6}
    Dataset& Methods &$ACC_{D_{test}}$$\uparrow$ & $ACC_{D_{f}}$ & $\Delta$$\downarrow$ & AD$\downarrow$  \\       
\midrule
\multirow{6}{*}{\makecell{ResNet-18}}&Gold Model         &81.03 &81.29 &0     & 0      \\
&Fine-tuning     &80.61 &100   &18.71  & 0.56   \\
&NegGrad &80.27 &23.94 &57.37  & 0.72   \\
&RandLabel    &80.09 &14.35 &66.94  & 0.83   \\
&Bad Teacher     &78.45 &60.64 &19.35  & 0.41   \\
&Ours            &79.55 &89.29 &7.94   & 0.28   \\
\midrule
\multirow{6}{*}{\makecell{ViT}}&Gold Model         &82.00 &84.71 &0     & 0      \\
&Fine-tuning     &83.53 &100.0 &15.29 & 0.50 \\
&NegGrad &80.50 &39.20 &45.51 & 0.75  \\
&RandLabel    &83.10 &0     &84.71 & 0.74  \\
&Bad Teacher     &81.19 &81.47 &8.64  & 0.25 \\
&Ours            &82.00 &88.76 &6.23  & 0.19 \\

\bottomrule
\end{tabular}
\label{main_result2}
\vspace{-3mm}
\end{table}

\section{Evaluation}
\subsection{Experimental setting} \label{es}
\noindent\textbf{Datasets \& Models.}
We evaluate our approach using three public image datasets: CIFAR-10, CIFAR-100 \cite{krizhevsky2009learning} and VGGFaces2~\cite{cao2018vggface2}. CIFAR-10 comprises 10 classes, and we perform data forgetting evaluations for each class independently. For CIFAR-100, we randomly select 17 out of 100 classes for evaluation due to space constraints. Since we emphasize on forgetting only a subset of one class, we randomly select $100$ images as $D_f$. 
For VGGFaces2, we randomly select 10 out of 100 celebrities as the forgetting class, and then randomly choose $50$ facial images to comprise $D_f$ for each forgetting class.
%Besides, we assess the performance when forgetting a larger amount of data, \emph{i.e.,} by increasing the size of $D_f$ to $500$, $1000$, $2000$, $3000$ and $4000$. 

Three common image classification neural networks are employed for evaluation: ResNet-18 \cite{he2016deep}, AllCNN \cite{springenberg2014striving}, and Vision Transformer~\cite{lee2021vision}.

\noindent\textbf{Baseline Methods.}
We compare our approach against four machine unlearning methods:
(1)\emph{ Negative Gradient} \cite{golatkar2020eternal}:
fine-tune the original model on $D$ by increasing the loss for samples in $D_f$, which is the common surrogate objective adopted by most unlearning methods. 
(2)\emph{ Fine-tuning}: fine-tune the model on $D_r$ using a slightly large learning rate. This is analogous to catastrophic forgetting, as fine-tuning without $D_f$ may cause the model to forget $D_f$.
(3)\emph{ Random Labeling} \cite{graves2021amnesiac}: fine-tune the model on
$D$ by assigning random labels to samples in $D_f$, causing those samples to receive a random gradient.
(4)\emph{ Bad Teacher} \cite{chundawat2023can}: it explores the utility of competent and incompetent teachers
in a student-teacher framework to induce forgetfulness. Note that this work is closely related to our approach since it also emphasizes alignments in data forgetting.

\begin{table*}[h]
\caption{Comparison on \textbf{VGGFaces2} dataset. Regarding $M_o$, its average accuracy over 10 classes is $87.01\%$.}
\scriptsize
 \centering
 \resizebox{1.02\textwidth}{!}{
 \setlength{\tabcolsep}{0.3em}%
  \begin{tabular}{c|c c|c c|c c|c c|c c|c c}
  \toprule
  \multirow{2}{*}{Removal Class} & 
  \multicolumn{2}{c|}{Gold model} & 
  \multicolumn{2}{c|}{Fine-tuning} &
  \multicolumn{2}{c|}{Negative Gradient} & 
  \multicolumn{2}{c|}{Random Labeling} &   
  \multicolumn{2}{c|}{Bad Teacher} & 
  \multicolumn{2}{c}{Ours} \\
  & ACC$^g_{D_{test}}$ & ACC$^g_{D_{f}}$ 
  & ACC$_{D_{test}}$ & ACC$_{D_{f}}$
  & ACC$_{D_{test}}$ & ACC$_{D_{f}}$; $\Delta$
  & ACC$_{D_{test}}$ & ACC$_{D_{f}}$
  & ACC$_{D_{test}}$ & ACC$_{D_{f}}$; $\Delta$
  & ACC$_{D_{test}}$ & ACC$_{D_{f}}$; $\Delta$  \\
  \midrule
    Class 0   & 85.81  & 94  & 87.09 &100  & 85.91 & 63; 31           &84.26 &0 &84.51  &91; \enspace3  &85.86     &90; \enspace4 \\
    Class 1   & 85.31  & 83  & 86.89 & 100 & 85.72 & 67; 14    & 85.52 &0 &83.31  &72; \enspace9  &86.49   &82; \enspace1 \\ 
    Class 2   & 84.61 & 91  & 86.60 & 100 & 85.21 & 60; 38  & 85.79 &1 &84.21  & 80; 11    & 86.16  & 92; \enspace1 \\ 
    Class 3   & 85.31 & 87  & 86.50 & 100 & 85.11 & 58; 39            & 84.50 &0 &84.61  &89; \enspace2    & 86.59 & 85; \enspace2\\ 
    Class 4   & 85.11 & 97  & 86.70 & 100 & 85.50 & 66; 30             & 85.89 &0 &85.91  &94; \enspace4    &85.50    &94; \enspace3 \\ 
    Class 5   & 85.02 & 98  & 86.79 & 100 & 85.01 & 56; 40            & 85.95 &0 &85.21  &85; 13    &86.15    &91; \enspace7 \\
    Class 6   & 85.71 & 95 & 86.70 & 100 & 84.64 & 53; 31            & 84.84 &0&85.81  &86;  \enspace9  &86.67    &94; \enspace1 \\
    Class 7   & 85.41 & 81  & 86.79 & 100 & 85.38 & 63; 16     & 84.61 &0 &84.61 &68; 13   &87.28     &90; \enspace9 \\
    Class 8   & 85.42 & 87  &86.89  & 100 & 84.93 & 48; 39     & 85.15 &0 &83.21  &55; 33   &87.08     &87; \enspace0 \\
    Class 9   &84.82  & 94   & 86.60 & 100 &85.16  &52; 42   & 83.85 &0 &83.61  &84; 10   &86.38    &96; \enspace2 \\
    Avg   &85.25  &90.7  &86.76  &100  &85.23  & 60.5; $\textbf{30.2}$ &85.01 &0.1  &84.50  &80.4; $\textbf{10.7}$  &86.41 &90.1; $\textbf{3}$ \\
  \bottomrule
  \end{tabular}
}
%\vspace{-5mm}
\label{vgg_100} 
\end{table*}

\noindent\textbf{Evaluation Metrics.}
We assess the alignment quality of unlearning methods using two metrics: 
(1) \emph{Accuracy on $D_f$ and $D_{test}$}: Ideally, the unlearned model is expected to achieve the same accuracy as the gold model. Let $ACC_{D_f}$ and $ACC^g_{D_f}$ denote the accuracy of the unlearned model and the gold model on $D_f$. Thus, the differences between them, $\Delta = |ACC_{D_f} - ACC^g_{D_f}|$, can be considered as the measure of alignment quality. Additionally, accuracy on $D_{test}$ is employed to assess whether the model accuracy is compromised after data forgetting. 
(2) \emph{Activation Distance}: This metric calculates the average L2-distance between the activation values of the unlearned model and the gold model on $D_f$. A smaller activation distance indicates a better alignment between two models.

\subsection{Main Results}
%For the distance feature, we measure the average distance of the five nearest data in the same class..
We compare our approach with four state-of-the-art unlearning methods. Table.\ref{main_result} presents the comparison on the CIFAR-10 dataset. When adopting the ResNet-18 neural network, the original model attains an average accuracy of $85.37\%$ across 10 classes. When adopting the AllCNN neural network, the original model has an average accuracy of $86.56\%$. In this evaluation, we conduct the data forgetting experiment for each class independently. For each class, $100$ samples are randomly selected as $D_f$. Due to space constraints, the class-wise results are shown in supplementary material, while Table.\ref{main_result} illustrates the average results across 10 classes.

\noindent\textbf{Accuracy on $D_f$ and $D_{test}$:}
We provide the accuracy of the gold model on both $D_{test}$ and $D_f$ (\emph{i.e.,} $ACC^g_{D_{test}}$ and $ACC^g_{D_f}$). For each method, we provide the $ACC_{D_{test}}$, $ACC_{D_f}$, and $\Delta = |ACC_{D_f}-ACC^g_{D_f}|$. The closer the accuracy between the gold model and the unlearned model (\emph{i.e.,} the smaller the $\Delta$ is), the better the alignment is achieved. 
In addition, the accuracy on $D_{test}$ is employed to assess whether the model's normal performance is compromised after data forgetting. The higher the accuracy on $D_{test}$ is, the better the normal performance is maintained.

From Table.\ref{main_result}, it's evident that the Fine-tuning method cannot effectively accomplish data forgetting, as the $ACC_{D_f}$ remains almost at $100\%$. It implies that Fine-tuning can only forget an entire class but struggles to forget a subset of a class. In contrast, the Random Labeling method tends to misclassify all samples in $D_f$, resulting in $ACC_{D_f}=0\%$. The Negative Gradient method performs better than Fine-tuning and Random Labeling, with an average difference of $\Delta_{NG}=76.3\%$. Due to explicitly addressing the alignment issue, the Bad Teacher method makes significant progress, achieving a remarkable average difference of $\Delta_{BadT}=8\%$. Nevertheless, our approach outperforms the Bad Teacher method in terms of both $\Delta$ and $ACC_{D_{test}}$. Particularly, we improve the average difference $\Delta$ from $8\%$ to $4.2\%$.

The comparison on the CIFAR-100 is shown in Table.\ref{main_result2}. We draw similar conclusions as with CIFAR-10. The Random Labeling method no longer misclassifies all samples in $D_f$, with an average $ACC_{D_f}=14.35\%$. Our approach demonstrates a greater advantage than the Bad Teacher method, with $\Delta_{ours}=7.94\%$ against $\Delta_{BadT}=19.35\%$. 
Note that we did not present results for the ALLCNN network, as it failed to classify 100 classes in the CIFAR-100 dataset.
%\vspace{-0.5em}
%\subsubsection{Network Architecture}
%\subsubsection{More Results}

The comparison on the VGGFaces2 dataset is presented in Table.\ref{vgg_100}. 
%Additionally, the comparison for the original model adopting AllCNN~\cite{springenberg2014striving} and ViT~\cite{lee2021vision} neural networks are presented in Table.\ref{} and Table.\ref{}. 
Similarly, our approach consistently outperforms all other methods, achieving the average $\Delta=3\%$.
% for the two network architectures.  

\noindent\textbf{Activation Distance:}
%\noindent\textbf{Activation Distance.}
Activation distance \cite{golatkar2021mixed} is another effective metric for evaluating the alignment in unlearning. A smaller activation distance indicates a higher similarity between the unlearned model and the gold model. 

\begin{table*}[t]
\caption{Ablation study on three discriminative features. We use the accuracy of predicted generalization-labels $ACC_{gl}$ as the metric to measure the quality of alignment.}
\scriptsize
\centering
\setlength{\tabcolsep}{0.3em}%
\begin{tabular}{c|c c | c c |c c |c c}
\toprule
\multirow{2}{*}{Forgetting Class} & 
\multicolumn{2}{c|}{DF+AF+CF (Ours)}  & 
\multicolumn{2}{c|}{DF}& 
\multicolumn{2}{c|}{AF}& 
\multicolumn{2}{c}{CF}\\
& $ACC_{D_{test}}$$\uparrow$ & $ACC_{gl}$$\uparrow$
& $ACC_{D_{test}}$$\uparrow$ & $ACC_{gl}$$\uparrow$
& $ACC_{D_{test}}$$\uparrow$ & $ACC_{gl}$$\uparrow$
& $ACC_{D_{test}}$$\uparrow$ & $ACC_{gl}$$\uparrow$
\\
\midrule
  Class 0 & 90.4 & \textbf{92} &89.2 &92 &90.1 &92        & 89.2& 91\\
  Class 1 & 94.4 & \textbf{100} &93.8 &92 &93.7 &97       & 94.3& 100\\
  Class 2 & 79.6 & \textbf{84} &72.6 &83 &71.1 &85        &75.1 &82\\
  Class 3 & 76.9 & 79 &69.2 &72 &65.5 &69        &76.6 &\textbf{81}\\
  Class 4 & 83.3 & \textbf{85} &80.3 &80 &82.3 &82        &83.7 &82\\
  Class 5 & 82.9 & 82 &78.5 &77 &77.0 &74        &81.5 &\textbf{86}\\
  Class 6 & 91.7 & 89 &88.2 &85 &89.3 &82        &91.0 &\textbf{90}\\
  Class 7 & 90.8 & \textbf{87} &91.1 &86&91.7 &86         &91.7 &84\\
  Class 8 & 94.7 & \textbf{92} &92.9 &90 &82.7 &89        &94.7 &91 \\
  Class 9 & 93.6 & \textbf{98} &92.0 &94 &92.0 &95        &93.2&96\\
\bottomrule
\end{tabular}
\label{classifier} 
\vspace{-1.0em}
\end{table*}

%\subsubsection{Comparison with Different DNNs}
In our experiments, three different neural network architectures are evaluated, including ResNet-18~\cite{he2016deep}, AllCNN~\cite{springenberg2014striving} and ViT~\cite{lee2021vision}.
From Table.\ref{main_result} and \ref{main_result2}, we found that the performance of unlearning methods is less influenced by the neural network architecture. Specifically, the Fine-tuning method maintains $ACC_{D_f}=100\%$, whereas in contrast, the Random Labeling method results in almost $ACC_{D_f}=0\%$. Both methods fail to achieve effective data forgetting, regardless of the employed network architecture. We did not present the results for AllCNN on CIFAR-100, as AllCNN is unable to effectively handle a 100-class classification problem.

The Negative Gradient method exhibits only a slight alignment effect, achieving an average difference $\Delta_{NG}^{AllCNN}=84.3\%$ on AllCNN and $\Delta_{NG}^{ViT}=23.9\%$ on ViT, respectively. In contrast, the Bad Teacher method makes more progress, achieving a remarkable average difference $\Delta_{BadT}^{AllCNN}=5.8\%$ on AllCNN and $\Delta_{BadT}^{ViT}=5.3\%$ on ViT, respectively. Nevertheless, our method still outperforms the Bad Teacher. Particularly, we improve the average difference $\Delta$ from $5.8\%$ to $4.9\%$ on AllCNN and  $\Delta$ from $5.3\%$ to $3.3\%$ on ViT, respectively.

%\subsection{Discussion}

\subsection{Ablation Study}
%\noindent\textbf{Ablation Study.}
In our approach, three discriminative features are proposed and combined to distinguish hard samples from easy samples. We will evaluate their individual contributions through an ablation study in this section. Since these features are developed to predict the generalization-labels on $D_f$, we use the prediction accuracy as the metric. Note that the prediction accuracy $ACC_{gl}$ is positively related to the quality of alignment (\emph{i.e.,} $\Delta$, AD, and ASR). The higher the accuracy, the better the alignment.
Table.\ref{classifier} shows the experimental results on CIFAR-10, where the original model adopts ResNet-18.
From Table.\ref{classifier}, the Curriculum-learning-loss Feature (CF) is slightly more discriminative than the Distance Feature (DF) and Adversarial-attacking Feature (AF). Nonetheless, the three features are complementary to each other.

\section{Conclusion}
This paper aims at the task of data forgetting, emphasizing the alignment in data forgetting. We introduce a Twin Machine Unlearning approach, where a twin unlearning problem is constructed and leveraged to solve the original problem. Furthermore, three discriminative features are devised by employing adversarial attack and curriculum learning strategies. Extensive empirical experiments show that our approach significantly improve the alignment in data forgetting.

\bibliographystyle{IEEEbib}
\bibliography{icme2025references}
% \bibliography{iclr2025_conference}

\end{document}

% 论文名称
\title{Supplementary Material: Towards Aligned Data Forgetting via Twin Machine Unlearning}

% 双盲，匿名
\author{ICME submission 1754}

% \author{\IEEEauthorblockN{Zhenxing Niu}
% \IEEEauthorblockA{\textit{Xidian University} \\
% Xian, China \\
% zxniu@xidian.edu.cn}\\
% %\and

% \IEEEauthorblockN{Yuhang Wang}
% \IEEEauthorblockA{\textit{Xidian University} \\
% Xian, China \\
% 24031212230@stu.xidian.edu.cn}

% \and

% \IEEEauthorblockN{Jiaqi Wang}
% \IEEEauthorblockA{\textit{Xidian University} \\
% Xian, China \\
% 22031212458@stu.xidian.edu.cn}\\

% \IEEEauthorblockN{Maoguo Gong}
% \IEEEauthorblockA{\textit{Xidian University} \\
% Xian, China \\
% gong@ieee.org}

% \and

% \IEEEauthorblockN{Haoxuan Ji}
% \IEEEauthorblockA{\textit{Xi'an Jiaotong University} \\
% Xian, China \\
% xjtujhx@stu.xjtu.edu.cn}\\

% \IEEEauthorblockN{Quan Wang}
% \IEEEauthorblockA{\textit{Xidian University}\\
% Xian, China \\
% qwang@xidian.edu.cn}
% }

\maketitle

\section{Class-wise Results for Data Forgetting}

We conduct the data forgetting experiment for each class in CIFAR-10 and CIFAR-100 independently, as depicted in each row of Table.\ref{tbc10}, Table.\ref{tbc100}, Table.\ref{tb1}, Table.\ref{tb2} and Table.\ref{tb3}. For each class, 100 samples are randomly selected as $D_f$.

\begin{table*}[htp]
\caption{Comparison with \textbf{ResNet-18} on \textbf{CIFAR-10} dataset. We conduct the data forgetting experiment for each class in CIFAR-10 independently, where $100$ images are randomly selected as $D_f$.}
\scriptsize
 \centering
 \resizebox{\textwidth}{!}{
 \setlength{\tabcolsep}{0.15em}%
  \begin{tabular}{c|c c|c c|c c|c c|c c|c c}
  \toprule
  \multirow{2}{*}{Removal Class} & 
  \multicolumn{2}{c|}{Gold model} & 
  \multicolumn{2}{c|}{Fine-tuning} &
  \multicolumn{2}{c|}{Negative Gradient} & 
  \multicolumn{2}{c|}{Random Labeling} &   
  \multicolumn{2}{c|}{Bad Teacher} & 
  \multicolumn{2}{c}{Ours} \\

  & ACC_{D_{test}} & ACC_{D_{f}} 
  & ACC_{D_{test}} & ACC_{D_{f}}
  & ACC_{D_{test}} & ACC_{D_{f}}; $ \Delta$
  & ACC_{D_{test}} & ACC_{D_{f}}
  & ACC_{D_{test}} & ACC_{D_{f}}; $ \Delta$
  & ACC_{D_{test}} & ACC_{D_{f}}; $ \Delta$  \\
  
  \midrule
    Class 0  & 85.61 & 92  & 85.08& 100 &85.29 &\enspace8; 84    &85.49 &0  &83.98 & 95; \enspace3  &84.60 &96; \enspace4      \\
    Class 1  & 85.27 & 97 & 85.07 & 100 &85.31  &11; 86    &85.75 &0  &80.45 & 91; \enspace6  &84.76 &98; \enspace1    \\ 
    Class 2  & 85.66 & 84 & 84.93 & 100 &85.03 &11; 73    &85.03 &0  &84.27 & 73; 11 &84.65 &85; \enspace1   \\ 
    Class 3  & 84.8 & 74 & 84.97 & 100 &85.09  &\enspace9; 65     &85.58 &0  &80.51 & 80; \enspace6  &85.07 &72; \enspace2         \\ 
    Class 4  & 85.24 & 82 & 84.94 & 100 &85.29 &15; 67    &85.22 &0  &81.96 & 94; 12 &84.86 &92; 10    \\ 
    Class 5  & 85.49 & 76 & 85.03 & 100 &85.06 &13; 63    & 85.37 &0  &82.69 & 98; 22 &84.43 &81; \enspace5     \\
    Class 6  & 85.43 & 88 & 84.98 & 100 &85.31 &\enspace9; 79    &85.62 &0  &82.64 & 98; 10 &85.02 &85; \enspace3   \\
    Class 7  & 84.97 & 87 & 84.90 & 100 &85.18 &\enspace9; 78    &85.63  &0  &82.87 & 92; \enspace5  &85.10 &99; 12   \\
    Class 8  & 85.13 & 94 & 85.02 & 100 &81.34 &\enspace2; 92    &85.42 &0  &83.11 & 97; \enspace3  &85.04 &96; \enspace2   \\
    Class 9  & 85.53 & 95 & 85.12 & 100 &85.25 &19; 76    &85.32 &0  &82.93 & 97; \enspace2  &84.49 &97; \enspace2     \\
    Avg   &85.31 &86.9 &85.06  &100  &84.81  &10.6; $\textbf{76.3}$ &85.43  &0  &82.37  &91.5; $\textbf{8}$  &84.86 &90.1; $\textbf{4.2}$  \\
  \bottomrule
  \end{tabular}
}
\label{tbc10} 
\end{table*}

\begin{table*}[htpb]
 \caption{Comparison with \textbf{ResNet-18} on \textbf{CIFAR-100} dataset. We randomly sample 17 classes for evaluation, where $100$ images are randomly selected as $D_f$.}
\scriptsize
 \centering
 \resizebox{\textwidth}{!}{
 \setlength{\tabcolsep}{0.1em}%
  \begin{tabular}{c|c c|c c|c c|c c|c c|c c}
  \toprule
  \multirow{2}{*}{Removal Class} & 
  \multicolumn{2}{c|}{Gold model} & 
  \multicolumn{2}{c|}{Fine-tuning} &
  \multicolumn{2}{c|}{Negative Gradient} & 
  \multicolumn{2}{c|}{Random Labeling} &   
  \multicolumn{2}{c|}{Bad Teacher} & 
  \multicolumn{2}{c}{Ours} \\
  & ACC_{D_{test}} & ACC_{D_{f}} 
  & ACC_{D_{test}} & ACC_{D_{f}}
  & ACC_{D_{test}} & ACC_{D_{f}} ; $\Delta$
  & ACC_{D_{test}} & ACC_{D_{f}} ; $\Delta$
  & ACC_{D_{test}} & ACC_{D_{f}} ; $ \Delta$
  & ACC_{D_{test}} & ACC_{D_{f}} ; $ \Delta$\\
  \midrule
    road       & 81.07 & 92 &80.61 &100 &80.62  &30; 62 &80.00 &52; 89 &78.11 &91; \enspace1 &78.97 &99; \enspace7 \\
    turtle     & 81.04 & 79 &80.65 &100        &80.44 &25; 54 &80.24 &\enspace7; 72 &78.91 &25; 54 &79.09 &84; \enspace5  \\ 
    chimpanzee & 81.48 & 90 &80.64 &100        &80.82 &28; 62 &80.11 &\enspace8; 82 &77.55 &70; 20 &79.69 &96; \enspace6   \\ 
    orchid     & 81.15 & 86 &80.60 &100        &80.48 &30; 56 &80.17 &17; 69 &78.64 &80; \enspace6 &79.68 &99; 13 \\ 
    rabbit     & 81.25 & 70 &80.54 &100        &80.31 &17; 53 &79.99 &\enspace1; 69 &78.15 &23; 47 &79.17 &91; 21  \\ 
    forest     & 81.33 & 72 &80.64 &100        &80.31 &22; 50 &79.80 &17; 55 &79.24 &74; \enspace2 &79.71 &89; 17  \\
    possum     & 80.40  & 76 &80.57 &100 &79.82 &\enspace1; 75 &80.02 &11; 65 &77.96 &59; 17 &79.84 &84; \enspace8  \\
    fox        & 81.18 & 89 &80.65 &100 &80.38 &23; 66 &79.94 &14; 75 &78.43 &64; 25 &80.21 &91; \enspace2 \\
    house      & 81.76 & 85 &80.47 &100  &80.67 &24; 61 &79.79 &\enspace9; 76 &78.11 &74; 11 &79.97 &85; \enspace0   \\
    mushroom   & 81.04 & 79 &80.67 &100  &80.53 &21; 58 &79.98 &10; 69 &78.99 &64; 15 &79.4  &96;  17  \\ 
    chair      &81.05  &90  &80.50 &100 &80.74 &39; 51 &79.96 &34; 56 &78.37 &85; \enspace5 &79.75 &94; \enspace4   \\
    tiger      &80.79  &86  &80.74 &100 &80.51 &30; 56 &80.21 &\enspace6; 80 &78.58 &59; 27 &78.45 &96; 10   \\
    snail      &81.29  &82  &80.56 &100 &70.10 &15; 67 &80.21 &\enspace4; 78 &78.21 &51; 31 &79.77 &84; \enspace2   \\
    worm       &80.92  &90  &80.72 &100 &80.34 &41; 49 &80.28 &26; 64&78.56 &87; \enspace3 &80.10  &90; \enspace0  \\
    beetle     &81.16  &82  &80.61 &100 &80.66 &27; 55 &80.06 &17; 65&78.74 &64; 18 &79.42 &84; \enspace2   \\
    beaver     &81.21  &54  &80.67 &100 &80.32 &\enspace6; 48 &80.31 &\enspace5; 49 &78.33 &27; 27  &79.36 &74; 20  \\
    bed        &79.36  &80  &80.54 &100 &80.66 &28; 52 &80.41 &\enspace6; 68 &78.8  &60; 20   &79.85 &81; \enspace1   \\
    Avg   &81.03 &81.29  &80.61 &100 &80.27 &23.94; $\textbf{57.37}$ &80.09 &14.35; $\textbf{66.94}$ &78.45 &60.64; $\textbf{19.35}$ &79.55 &89.29; $\textbf{7.94}$ \\
  \bottomrule
  \end{tabular}
}
\vspace{-4mm}
\label{tbc100} 
\end{table*}

\begin{table*}[h]
\caption{Comparison with \textbf{AllCNN} on \textbf{CIFAR-10} dataset. Regarding $M_o$, its average accuracy over 10 classes is $87.21\%$.}
\scriptsize
 \centering
 \resizebox{\textwidth}{!}{
 \setlength{\tabcolsep}{0.15em}%
  \begin{tabular}{c|c c|c c|c c|c c|c c|c c}
  \toprule
  \multirow{2}{*}{Removal Class} & 
  \multicolumn{2}{c|}{Gold model} & 
  \multicolumn{2}{c|}{Fine-tuning} &
  \multicolumn{2}{c|}{Negative Gradient} & 
  \multicolumn{2}{c|}{Random Labeling} &   
  \multicolumn{2}{c|}{Bad Teacher} & 
  \multicolumn{2}{c}{Ours} \\
  
  & ACC^g_{D_{test}} & ACC^g_{D_{f}} 
  & ACC_{D_{test}} & ACC_{D_{f}} 
  & ACC_{D_{test}} & ACC_{D_{f}}; $\Delta$
  & ACC_{D_{test}} & ACC_{D_{f}} 
  & ACC_{D_{test}} & ACC_{D_{f}}; $\Delta$
  & ACC_{D_{test}} & ACC_{D_{f}}; $\Delta$  \\
  
  \midrule
    Class 0  & 87.0\enspace  & 94 & 87.04 & 100   & 86.69 & 3; 91  & 86.96 &0  & 86.09& 86; 8   &87.03 &89; 5    \\
    Class 1  & 86.58 & 97 & 87.49 & 100   & 86.86 & 2; 95  & 86.99 &0 & 86.32& 92; 5  &87.43 &98; 1   \\ 
    Class 2  & 87.09 & 82 & 87.42 & 100  & 86.68  & 5; 77   & 86.53 &0 & 85.71& 75; 7   & 87.06  &87; 5   \\ 
    Class 3  & 86.22 & 78 & 87.38 & 100  & 86.85 & 2; 76  & 86.62 & 0 & 86.54& 80; 2  & 87.51  &69; 9   \\ 
    Class 4  & 86.4\enspace  & 87 & 87.39 & 100  & 86.89 & 3; 84  & 87.72 & 0 &84.45& 96; 9    &86.46 &92; 5   \\ 
    Class 5  & 86.36 & 79 & 87.23 & 100  & 86.75 & 5; 74  & 86.52 & 0 & 85.98& 86; 7 & 87.37  &83; 4    \\
    Class 6  & 86.46 & 87 & 87.42 & 100  & 87.00 & 1; 86  & 87.39 & 0 & 86.44& 91; 4  &86.86 &89; 2   \\
    Class 7  & 86.24 & 82 & 87.31 & 100  & 86.79 & 1; 81 & 87.24  & 0 &86.53& 91; 9   & 87.24  &89; 7  \\
    Class 8  & 87.0\enspace & 91 & 87.65  &100   &87.22  & 6; 85  & 86.95 & 0 &84.51 &90; 1   &86.84 &87; 4     \\
    Class 9  & 86.54 & 87 & 87.16 & 100  & 86.93 & 5; 82  & 86.02 & 0 & 87.05 & 93; 6  &87.05 &94; 7    \\
    Avg    &86.58  &86.4  &87.34  &100   &86.87  &3.3; $\textbf{83.1}$ &86.89  &0   &85.96 &88.0; $\textbf{5.8}$    &86.78 &87.7; $\textbf{4.9}$  \\
  \bottomrule
  \end{tabular}
}
\vspace{-4mm}
\label{tb1} 
\end{table*}

\begin{table*}[h]
\caption{Comparison with the \textbf{ViT} model on \textbf{CIFAR-10} dataset. Regarding $M_o$, its average accuracy over 10 classes is $84.45\%$.}
\scriptsize
 \centering
 \resizebox{\textwidth}{!}{
 \setlength{\tabcolsep}{0.15em}%
  \begin{tabular}{c|c c|c c|c c|c c|c c|c c}
  \toprule
  \multirow{2}{*}{Removal Class} & 
  \multicolumn{2}{c|}{Gold model} & 
  \multicolumn{2}{c|}{Fine-tuning} &
  \multicolumn{2}{c|}{Negative Gradient} & 
  \multicolumn{2}{c|}{Random Labeling} &   
  \multicolumn{2}{c|}{Bad Teacher} & 
  \multicolumn{2}{c}{Ours} \\

  & ACC^g_{D_{test}} & ACC^g_{D_{f}} 
  & ACC_{D_{test}} & ACC_{D_{f}}
  & ACC_{D_{test}} & ACC_{D_{f}}; $\Delta$
  & ACC_{D_{test}} & ACC_{D_{f}}
  & ACC_{D_{test}} & ACC_{D_{f}}; $\Delta$
  & ACC_{D_{test}} & ACC_{D_{f}}; $\Delta$  \\
  
  \midrule
    Class 0  & 84.01 & 92 & 84.70  & 100 &84.46   &62; 30   &84.16 &4 &83.49 & 92; 0  &82.94 &88; 4      \\
    Class 1  & 84.26 & 96 & 84.71 & 100 &84.58 &76; 20   &84.23 &9 &83.47 & 87; 9  &83.37 &96; 0    \\ 
    Class 2  & 85.66 & 88 & 84.54 & 100&84.58  &60; 28   &84.33 &1 &82.90  & 82; 6  &83.97 & 91; 3   \\ 
    Class 3  & 84.80 & 70  & 84.67 & 100 &84.35 &41; 29   &84.17 &0 &83.02 & 87; 17 &84.16  &  74; 4    \\ 
    Class 4  & 85.24 & 87 & 84.41 & 100 &84.56 &51; 36   &84.30  &0 &83.19 & 87; 0  &83.85  &86; 1    \\ 
    Class 5  & 84.32 & 78 & 84.39 & 100&84.37  &54; 24   &84.13 &0 &83.01 & 86; 8  &83.67  &80; 2     \\
    Class 6  & 83.39 & 85 & 84.78 & 100&84.54  &71; 14   &84.14 &3 &83.38 & 93; 8  &83.85   &90; 5   \\
    Class 7  & 84.97 & 85 & 84.74 & 100&84.63  &66; 19   &84.29 &2 &82.51 & 85; 0  &83.85 &79; 6   \\
    Class 8  & 85.13 & 92 & 85.7  & 100&84.54   &72; 20   &84.28 &3 &83.37 & 90; 2  &83.65 &89; 3   \\
    Class 9  & 84.38 & 93 & 85.65 & 100&84.50  &74; 19   &84.08 &2 &83.41 & 90; 3  &83.62 &88; 5     \\
    Avg   &84.61 &  86.6& 84.83 &100  &84.51  &62.7; $\textbf{23.9}$ &84.21  &2.4 &83.05  &87.9; $\textbf{5.3}$  &84.07 &86.1; $\textbf{3.3}$  \\
  \bottomrule
  \end{tabular}
}
\vspace{-4mm}
\label{tb2} 
\end{table*}

\begin{table*}[h]
 \caption{Comparison with the \textbf{ViT} model on \textbf{CIFAR-100} dataset. We randomly sample 17 classes for evaluation. The average accuracy of $M_o$ over 100 classes is $83.04\%$.}
\scriptsize
 \centering
 \resizebox{\textwidth}{!}{
 \setlength{\tabcolsep}{0.15em}%
  \begin{tabular}{c|c c|c c|c c|c c|c c|c c}
  \toprule
  \multirow{2}{*}{Removal Class} & 
  \multicolumn{2}{c|}{Gold model} & 
  \multicolumn{2}{c|}{Fine-tuning} &
  \multicolumn{2}{c|}{Negative Gradient} & 
  \multicolumn{2}{c|}{Random Labeling} &   
  \multicolumn{2}{c|}{Bad Teacher} & 
  \multicolumn{2}{c}{Ours} \\
  
  & ACC^g_{D_{test}} & ACC^g_{D_{f}} 
  & ACC_{D_{test}} & ACC_{D_{f}}
  & ACC_{D_{test}} & ACC_{D_{f}}; $\Delta$
  & ACC_{D_{test}} & ACC_{D_{f}}
  & ACC_{D_{test}} & ACC_{D_{f}}; $\Delta$
  & ACC_{D_{test}} & ACC_{D_{f}}; $\Delta$  \\
  \midrule
    road       &81.97 & 95 &83.43 &100  &80.51 &41; 54   &83.04 &0  &81.80 &95; \enspace0 &82.13 &100; \enspace5 \\
    turtle     &81.88 & 80 &83.45 &100  &80.54  &22; 58 &80.54 &0  &82.00 &65; 15 &80.61&\enspace83; \enspace3 \\ 
    chimpanzee &82.26 & 96 &83.43 &100  &80.43 &28; 68 &83.52 &0   &81.30 &81; 15 &82.42 &100; \enspace4\\ 
    orchid     &81.70 & 93 &83.42 &100  &80.37 &25; 68   &83.08 &0   &80.59&72; 21 &81.96 &\enspace87; \enspace6 \\ 
    rabbit     &82.11 & 75 &83.46 &100  &80.44 &14; 63    &82.95 &0   &81.09&48; 27 &82.42 &\enspace89; 14  \\ 
    forest     &82.49 & 77 &83.45&100  &80.37 &24; 54  &83.20  &0  &80.93 &76; \enspace1  &82.20  &\enspace90; 13 \\
    possum     &82.39 & 78 &83.41&100  &80.73 &16; 62 &83.36  &0  &80.98 &57; 21  &82.38 &\enspace84; \enspace6  \\
    fox        &81.67 & 90 &83.43&100  &80.43 &24; 66  &83.11  &0  &80.55 &92; \enspace2  &82.19 &\enspace86; \enspace4  \\
    house      &82.26 & 86 &83.42&100  &80.59 &24; 59  &83.35  &0  &81.03 &83; \enspace3  &82.31 &\enspace67; 19    \\
    mushroom   &81.22 & 88 &83.45&100  &80.73 &20; 66 &83.27  &0  &81.43 &94; \enspace6  &82.43 &\enspace95; \enspace7    \\
    chair      &83.51& 88 &83.43&100  &80.67 &39; 49  &83.11  &0  &81.45 &95; \enspace7  &82.41 &\enspace94; \enspace6    \\ 
    tiger      &82.81& 87 &83.47&100  &80.54 &35; 52 &83.15  &0  &81.11 &94; \enspace7  &81.57 &\enspace84; \enspace3    \\
    snail      &83.79& 90 &83.41&100  &80.54 &18; 72  &82.19  &0  &80.80  &86; \enspace4  &82.4  &\enspace94; \enspace4   \\
    worm       &83.22& 92 &83.41&100  &80.61 &38; 54  &83.04  &0  &81.22 &96; \enspace4  &82.56 &\enspace99; \enspace7    \\
    beetle     &83.12& 80 &83.39&100  &80.01 &15; 65  &83.15  &0  &81.32 &92; 12  &81.6 &\enspace97; 17   \\
    beaver     &83.05& 67 &83.44&100  &80.37 &\enspace6; 61  &83.50  &0  &81.37 &71; \enspace4  &81.76 &\enspace71; \enspace4    \\
    bed        &83.30& 81 &83.44&100  &80.57 &25; 56  &83.25  &0  &81.29 &88; \enspace7  &81.7  &\enspace89; \enspace8   \\
    avg  &82.51  &84.71 &83.43 &100 &80.50 &39.2; $\textbf{45.51}$ &83.10 &0 &81.19 &81.47; $\textbf{8,64}$ &82.06 &88.76; $\textbf{6.23}$ \\
  \bottomrule
  \end{tabular}
}
\vspace{-4mm}
\label{tb3} 
\end{table*}